\documentclass[a4paper, 10pt, conference]{ieeeconf}      % Use this line for a4
\usepackage{FG2021}

\FGfinalcopy % *** Uncomment this line for the final submission

\usepackage{tabularx,booktabs}
\usepackage{lipsum}
\usepackage{graphicx}

\IEEEoverridecommandlockouts                              % This command is only
\usepackage{amsmath} % assumes amsmath package installed
\usepackage{amssymb}  % assumes amsmath package installed

\def\FGPaperID{****} % *** Enter the FG2021 Paper ID here
\def\B#1{\textbf{#1}}
\def\BB#1{[\B{#1}]}

\title{\LARGE \bf
Multi-Modal Learning for AU Detection Based on Multi-Head Fused Transformers
}

\author{\parbox{16cm}{\centering
    {\large Xiang Zhang and Lijun Yin}\\
    {\normalsize
    Department of Computer Science, State University of New York at Binghamton, NY, USA\\}}
}

\begin{document}

%%%%%%%%%%%%%%
% COPYRIGHT NOTICE - Uncomment correct version below
%
% The notices are from the FG 2021 LOA 
%
% Active is the "Others" option - see Case #4 in the instructions posted at: http://iab-rubric.org/fg2021
%
%%%%%%%%%%%%%%

% Case #1: For papers in which all authors are employed by the US government, the copyright notice is: 
%\IEEEoverridecommandlockouts\pubid{\makebox[\columnwidth]{U.S. Government work not protected by U.S. copyright \hfill}
%\hspace{\columnsep}\makebox[\columnwidth]{ }}

% Case #2: For papers in which all authors are employed by a Crown government (UK, Canada, and Australia), the copyright notice is:
%\IEEEoverridecommandlockouts\pubid{\makebox[\columnwidth]{978-1-6654-3176-7/21/\$31.00~\copyright{}2021 Crown \hfill}
%\hspace{\columnsep}\makebox[\columnwidth]{ }}

% Case #3: For papers in which all authors are employed by the European Union, the copyright notice is:
%\IEEEoverridecommandlockouts\pubid{\makebox[\columnwidth]{978-1-6654-3176-7/21/\$31.00~\copyright{}2021 European Union \hfill}
%\hspace{\columnsep}\makebox[\columnwidth]{ }}

% Case #4: For all other papers the copyright notice is:
\IEEEoverridecommandlockouts\pubid{\makebox[\columnwidth]{978-1-6654-3176-7/21/\$31.00~\copyright{}2021 IEEE \hfill}
\hspace{\columnsep}\makebox[\columnwidth]{ }}

\ifFGfinal
\thispagestyle{empty}
\pagestyle{empty}
\else
\author{Anonymous FG2021 submission\\ Paper ID \FGPaperID \\}
\pagestyle{plain}
\fi
\maketitle

%%%%%%%%%%%%%%%%%%%%%%%%%%%%%%%%%%%%%%%%%%%%%%%%%%%%%%%%%%%%%%%%%%%%%%%%%%%%%%%%
\begin{abstract}
Multi-modal learning has been intensified in recent years, especially for applications in facial analysis and action unit detection whilst there still exist two main challenges in terms of 1) relevant feature learning for representation and 2) efficient fusion for multi-modalities. Recently, there are a number of works have shown the effectiveness in utilizing the attention mechanism for AU detection, however, most of them are binding the region of interest (ROI) with features but rarely apply attention between features of each AU.
On the other hand, the transformer, which utilizes a more efficient self-attention mechanism, has been widely used in natural language processing and computer vision tasks but is not fully explored in AU detection tasks. 
In this paper, we propose a novel end-to-end Multi-Head Fused Transformer (MFT) method for AU detection, which learns AU encoding features representation from different modalities by transformer encoder and fuses modalities by another fusion transformer module.
Multi-head fusion attention is designed in the fusion transformer module for the effective fusion of multiple modalities.
Our approach is evaluated on two public multi-modal AU databases, BP4D, and BP4D+, and the results are superior to the state-of-the-art algorithms and baseline models.
We further analyze the performance of AU detection from different modalities.

\end{abstract}

%%%%%%%%%%%%%%%%%%%%%%%%%%%%%%%%%%%%%%%%%%%%%%%%%%%%%%%%%%%%%%%%%%%%%%%%%%%%%%%%
% In particular, we first fed 12 AU embedding features of each modality into a transformer encoder for AU features encoding.
% Afterward, cross-modality multi-head attention is designed in another transformer for the fusion of features from multiple modalities.

\section{INTRODUCTION}

Facial action units (AUs), defined by Facial Action Coding System (FACS) \cite{Ekman1997}, has been widely used for understanding human emotions. Different from prototypical facial expression recognition (FER), AUs describe facial muscle movements in facial expressions.

With the advance of the deep learning networks, deep features show much better generalization than traditional hand-crafted features like HoG, SIFT, etc.
Their successful applications on large-scale datasets also benefit the AUs study. Comparing with other tasks, there exist some practical limitations: (1) accurate AU annotations usually require intensive human labor and professional experience; (2) provocation of spontaneous facial expressions is very difficult; (3) unbalanced labels problem, e.g., \textit{AU6} (Cheek Raiser) is more likely to occur than average but \textit{AU2} (Outer Brow Raiser) is less.

Based on the RGB modality, several methods achieve the state-of-the-art performance in AU detection \cite{eac,jaa,niu_2019,dsin,srerl,sev}. 
Meanwhile, the attention mechanism has been adopted in recent AU detection works. Most of them utilized an attention mechanism to find the region of interest (ROI) of each AU.
However, prior knowledge, e.g., facial landmarks are required to predefine the attentions of AUs. Therefore, it's more efficient to utilize a self-attention technique for learning. 
Moreover, little work explores the attention among the AUs learning features but instead focuses on the ROI-based attention.
On the other hand, transformer, based on self-attention mechanisms, achieved significant improvements in not only natural language processing (NLP) \cite{transformer,bert,brown2020,raffel2019} but also computer vision (CV) tasks \cite{vit,detr,zhu2020,yang2020}.
Therefore, we propose a new transformer-based network for AU encoding to tackle the above-mentioned inefficiency problem.

Data and task based multi-modalities fusions take the advantage of the complementary information and increase the robustness \cite{taskonomy}. 
Meanwhile, AU analysis relies on the facial muscle movement, the RGB-only detection methods are insufficient for recognizing subtle changes.
Thanks to recently developed multi-modal databases \cite{wang2012,bp4d+,bp4d} on facial expressions (e.g. RGB, 3D mesh, thermal, and physiology signals in BP4D+ \cite{bp4d+}), we can broaden our views from classic RGB to various modalities and from 2D image to 3D space.
For instance, \textit{AU6} (cheek raiser) involves the deformation of \textit{Orbicularis oculi} and \textit{pars orbitalis} muscles, which shows subtle differences when observed in visible images.
However, due to the high density of points around the eyes region, better geometric changes can be characterized on 3D mesh.
Although some works utilized the additional information presented from multiple modalities in FER and AU detection \cite{li2015,li2017,amf}, two main challenges remain: 1) the model must learn the most relevant features for representation; 2) the model must be effectively when fusing two modalities.

\begin{figure*}[ht]  %%%%%%%%%%%%%  Figure 1  %%%%%%%%%%%%%%%%%
    \centering
    \includegraphics[width=0.95\textwidth]{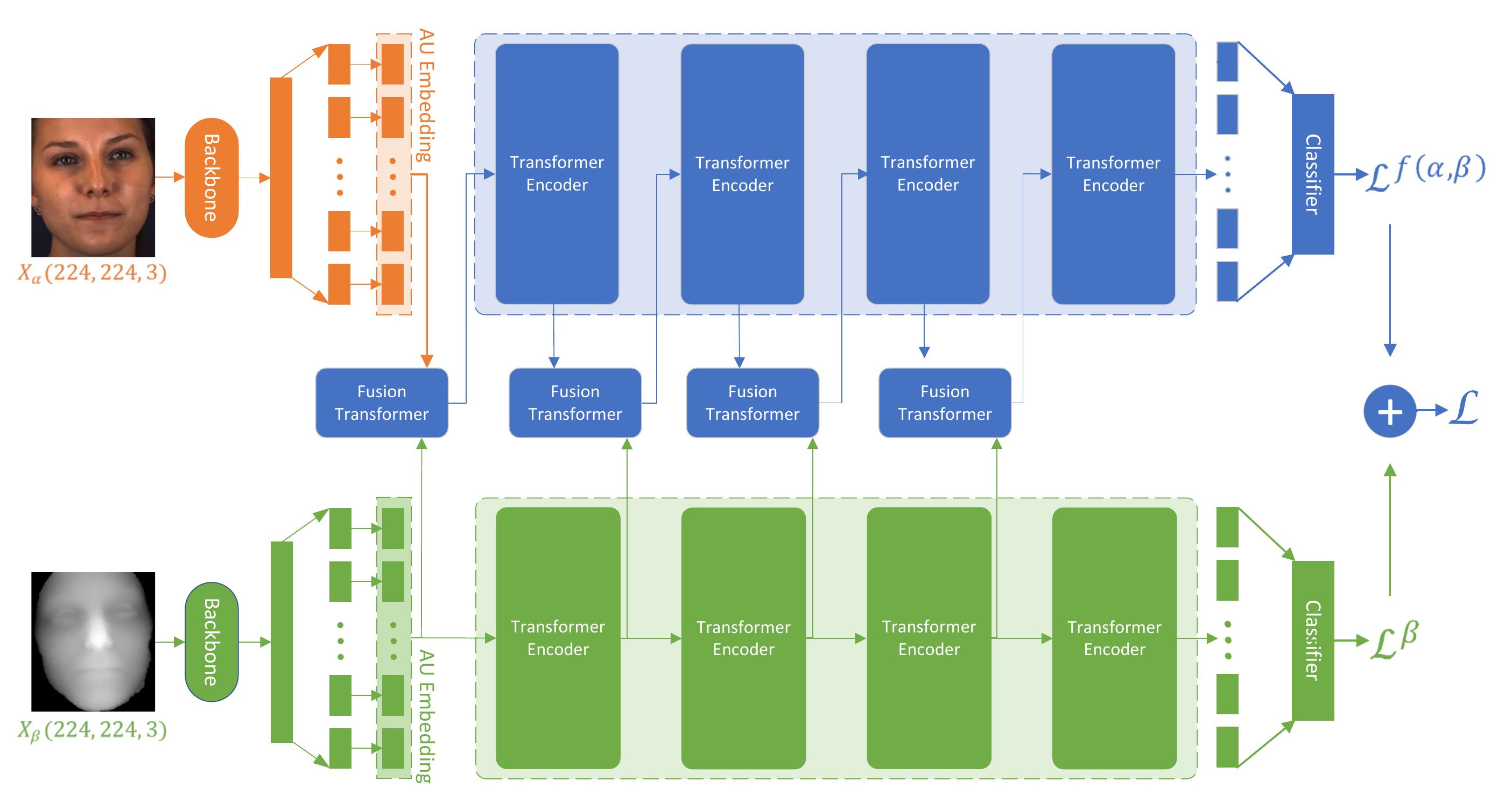}
    \caption{\textbf{Framework architecture.} The proposed framework consists of two pipelines with two modalities as inputs: $X_\alpha$, $X_\beta$. For example, modalities $\alpha$ and $\beta$ represent RGB and Depth respectively in the figure. 
    We design two core components in the network: Fusion Transformer (FT) module and AU Transformer Encoder(TE). 
    Two modalities are first extracted by a backbone network and then split the FC layer for each AU. 
    After obtaining AU embeddings of two modalities, they are fused through a FT module to acquire the fusion features $F(\alpha, \beta)$. Then following 4 TEs in each pipeline, $F(\alpha, \beta)$ and $\beta$ are encoded. Two features after each encoder of two pipelines are fused again by another FT module. Each pipeline ends with a classifier, and the final losses are combined from both. It is worth noting that $\alpha$ and $\beta$ can also represent Depth and RGB in turn, which means our fusion framework is sensitive to the fusion order.} 
    \label{fig:figure1}
\end{figure*}

Inspired by these vision transformer works \cite{vit,detr}, and multi-modal transformer works \cite{mult,transfuser}, we explore the methods that are utilized transformer architecture for both feature representation and fusion. 
In this paper, we study how multi-modal data help AUs analysis and present a novel Multi-Head Fused Transformer (MFT) for AU detection. Fig. 1\ref{fig:figure1} illustrates the proposed network, which contains two principal points to notice. 
First, two pipelines in our model, each pipeline is fed with AU embeddings from each modality. 
Second, transformer-based fusion modules are applied for the fusion of two modalities along the pipelines.
We design the multi-head fusion attention module in the fusion transformer to fuse modalities effectively.
As a result, our model not only learns the relevant AU encoding feature but also effectively fuses two modalities. 
We evaluate our methods on two public datasets in terms of BP4D \cite{bp4d} and BP4D+ \cite{bp4d+}. 
Our transformer based encoder performs better than the state-of-the-art methods on RGB images, which demonstrates the effectiveness of the AU encoding.
Moreover, the results of the fusion experiments with RGB and Depth images also show our method outperforms the SOTA methods, which indicates the fusion transformer module works effectively.

The main contribution of this work lies in three-fold: 

\begin{enumerate}
   \item We present a novel transformer based framework for AU detection, which can learn AU encoding features effectively. 
   \item We design a transformer based fusion module for multi-modalities fusion.
   To the best of our knowledge, this is the first attempt to utilize \B{multi-head fusion attention} to fuse AU features of two modalities in a transformer.
   \item Extensive experiments demonstrate that our network outperforms the state-of-the-art works in AU detection by both RGB-only data and RGB-Depth fusion features. 
\end{enumerate}

\begin{figure}[ht]   %%%%%%%%%%%%%  Figure 2  %%%%%%%%%%%%%%%%%
    \centering
    \includegraphics[width=0.45\textwidth]{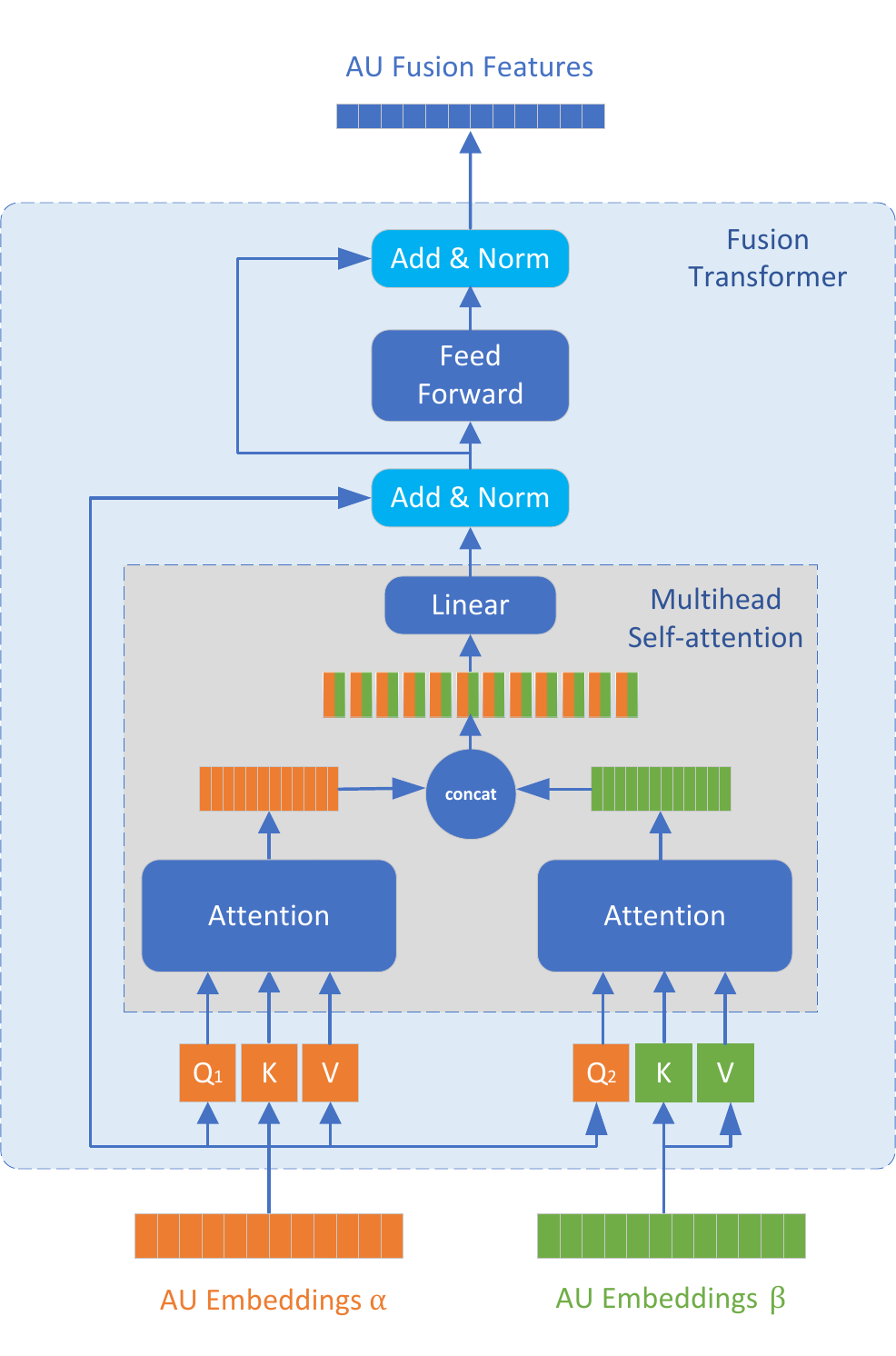}
    \caption{\textbf{Structure of Fusion Transformer module.} The module is based on a standard 2-heads attention transformer for modality $\alpha$ and modality $\beta$. 
    The difference is the attention of $\beta$ is calculated by $Q_\alpha$, $K_\beta$, $V_\beta$.
    The fusion features are then passed to AU Transformer encoders.}
    \label{fig:figure2}
\end{figure}

%%%%%%%%%%%%%%%%%%%%%%%%%%%%%%%%%%%%%%%%%%%%%%%%%%%%%%%%%%%%%%%%%%%%%%%%%%%%%%%%
\section{RELATE WORK}

We review the previous related works on AU detection and multi-modal deep learning.

\subsection{AU Detection}
Recent works on Facial Action Units detection achieved certain improvements by deep learning and attention mechanism. 
Zhao et al. \cite{drml} proposed a local feature learning approach that uniformly slices the first feature map as a region layer. 
Li et al. \cite{eac} proposed an EAC network, which predefines attentions based on AU-related landmarks to enhance and crop the ROI of AUs. 
Zhang et al. \cite{zhang2018} proposed an adversarial training method to maximize identity loss to focus on the AU-related features.
Shao et al. \cite{jaa} proposed the JAA-Net, which explores multi-scale region learning and adaptive attention learning.
Li et al. \cite{li2019} gave insight into the AU relation leaning by a Graph Neural Network (GCN) based model.
Niu et al. \cite{niu_2019} proposed a subject independent model to use the facial landmarks as person-specific shape regularization for AU detection.
These works with regional attention usually require prior information like facial landmarks. 
Recently, Yang et al. \cite{sev} proposed the SEV-Net that utilized the transformer for Intra-AU and Inter-AU attention learning from the AU semantic descriptions. However, the AU semantic descriptions are required as the prior knowledge.
Differ from the ROI-based attention, our transformer-based methods exploit the self-attention mechanism between AUs, which is no landmarks requirement.
Compare with the SEV-Net, our model only require the AU labels, and no AU semantic descriptions are needed. 

\subsection{Multi-modal Deep learning}

Multi-modal machine learning aims to build models that can process, correlate, and integrate information from multiple modalities \cite{multimodal}.
The success of deep learning has been a catalyst to solving
increasingly complex machine-learning problems, which
often involve multiple data modalities \cite{ramachandram2017}.
Multi-modal deep learning has been succeed used in a wide range of applications, e.g., human activity recognition \cite{mtut,neverova2015,emonets}, autonomous
driving \cite{nuscenes,cui2019,transfuser}, and image segmentation \cite{hu2018,Huang_2018}.

In the past decade, multi-modal learning has been studied on both facial expression recognition and action unit detection. 
Compare with visual images and audio \cite{emonets,2019avec} combination, especially dealing with videos, the potential of infrared, thermal, and 3D are not fully explored. 
Li et al. \cite{li2015,li2017} proposed approaches for facial expression recognition based on features fusion of 2D and 3D.
Irani et al. \cite{irani2015} proposed a method to utilize RGB-Thermal-Depth images for pain estimation. 
Lakshminarayana et al. \cite{lakshminarayana2019} explored the combination of physiological signals and RGB images to predict action units. 
Yang et al. \cite{amf} proposed a model called AMF, which included a feature scoring module to select the most relevant feature representations from different modalities.

Transformer \cite{transformer}, based on self-attention mechanism, was first applied to natural language processing tasks, then this framework has been extended to computer vision. 
For example, the ViT model, proposed by Dosovitskiy et al. \cite{vit}, applied a pure transformer to sequences of image patches, which achieved comparable results to CNNs on image classification.
Carion et al. \cite{detr} proposed the DETR framework, which combined a common CNN with a transformer encoder-decoder architecture.
Recently, researchers have explored multi-modal learning by transformers.
The most related works are called Multimodal Transformer (MulT) \cite{mult} and Multi-Modal Fusion Transformer (TransFuser) \cite{transfuser}, even though they target human multi-modal language analysis and multi-modal autonomous driving respectively.
In MulT, pairwise crossmodal transformers are designed, where the $Q$, and $KV$ are from different modalities. Then the outputs of pairwise crossmodal transformers are concatenated for fusion. 
In the TransFuser work, two modalities fusion occurred before the self-attention in a transformer, then was split after the self-attention. 
\B{Differ} from these two works, our method fuses the features of two modalities by the \B{multi-head fusion attention} module in a transformer.

%%%%%%%%%%%%%%%%%%%%%%%%%%%%%%%%%%%%%%%%%%%%%%%%%%%%%%%%%%%%%%%%%%%%%%%%%%%%%%%%
\section{PROPOSED METHOD}
The proposed Multi-Head Fused Transformer (MFT) is sketched in Fig. \ref{fig:figure1}.
Each data modality goes through its own backbone to extract features and then output the AU embeddings from the fully connected (FC) layer. 
There are two pipelines, and each contains 4 transformer-based AU encoders and a classifier in the end. 
One pipeline's input is the fusion feature from a fusion transformer, which is fed with the AU embedding features of fusion features, and the other pipeline encode the AU features from a single modality embeddings.
In the end, two losses from these two pipelines are combined together for training.

\subsection{AU Embeddings}
Two pre-trained ResNet-50 \cite{resnet} are employed as backbone encoders for RGB and Depth images.
The feature out from the fully connected layer is divided by each AU and then fed into an embedding layer for AU embedding.
The embedding layer in our model is similar to the standard transformer \cite{transformer}.
Since the AU encoder feature shares the same position with AU embedding, there is no need for position embeddings. 
Meanwhile, the stand layer normalization is applied to AU embedding as well.

\subsection{Transformer Encoder}

We apply the multi-layer transformer encoder to encode each AU embedding feature.
The Transformer encoder is formed by multiple alternating layers, where each layer consists of multiheaded self-attention and wise feed-forward network blocks.
Layernorm (LN) \cite{layernorm} is utilized before every block and residual connector after every block.
To summary, the transformer encoder is defined as:

\begin{equation}
X_A = LN(MSA(X)) + X
\end{equation}

\begin{equation}
X_B = LN(FFN(X_A)) + X_A
\end{equation}
where $X$ is the input of the Transformer layer, $X_B$ is the output of the Transformer layer, $LN$ is Layernorm and $MSA$ is multihead self-attention.

\subsection{Fusion Transformer}
The architect of Fusion Transformer is shown in Fig. \ref{fig:figure2}.
Two modalities $\alpha$ and $\beta$ are denoted $X_{\alpha} \in \mathbb{R}^{T \times D}$ and $X_{\beta} \in \mathbb{R}^{T \times D}$ respectively.
$T$ and $D$ are represented as AU class numbers and AU embedding feature dimensions, respectively.
Inspired by the Multimodal Transformer in \cite{mult}, we designed a two-heads self-attention for fusion, which is the core component in the fusion transformer.
We define the Querys, Keys, Values as following: $Q_{\alpha}^i = X_{\alpha}W_{Q_{\alpha}}^i$,
$K_{\alpha} = X_{\alpha}W_{K_{\alpha}}$, $K_{\beta} = X_{\beta} W_{K_{\beta}}$,
$V_{\alpha} = X_{\alpha}W_{V_{\alpha}}$, and $V_{\beta} = X_{\beta}W_{V_{\beta}}$,
where $W_{Q_{\alpha}}^i \in \mathbb{R}^{T \times D}, i \in \{1,2\}$, 
$W_{K_{\alpha}} \in \mathbb{R}^{T \times D}$,
$W_{K_{\beta}}  \in \mathbb{R}^{T \times D}$,
$W_{V_{\alpha}} \in \mathbb{R}^{T \times D}$,
$W_{V_{\beta}}  \in \mathbb{R}^{T \times D}$.
Then the fusion multi-head self-attention equation $f(X_\alpha, X_\beta)$ for $\alpha$ and $\beta$ is defined:

\begin{align} \label{eq:fusion}
f(X_\alpha, X_\beta) &= Att(Q_\alpha^1, K_\alpha, V_\alpha) \oplus Att(Q_\alpha^2, K_\beta, V_\beta) \nonumber\\
    &= \sigma(\frac{Q_\alpha^1K_\alpha^T}{\sqrt{d_k}})V_\alpha \oplus
       \sigma(\frac{Q_\alpha^2K_\beta^T}{\sqrt{d_k}})V_\beta \nonumber\\
    &= \sigma(\frac{X_{\alpha}W_{Q_{\alpha}}^1(X_{\alpha}W_{K_{\alpha}})^T}{\sqrt{d_k}})V_\alpha \oplus\\
    &\qquad{} \sigma(\frac{X_{\alpha}W_{Q_{\alpha}}^2(X_{\beta}W_{K_{\beta}})^T}{\sqrt{d_k}})V_\beta \nonumber
\end{align}

where $\sigma$ is $softmax$, $\oplus$ is concatenation, and $d_k$ is dimension of queries and keys.

The other components in the fusion transformer are the same as the standard transformer encoder we mentioned above.
Based on the attention of $Q_{\alpha}^2K_{\beta}V_{\beta}$, the fusion transformer enables the $\alpha$ modality to receive information from $\beta$ modality and then combines them.
It is worth noting that equation (\ref{eq:fusion}) can be easily extended to more modalities as following:
\begin{align}
    f(X_1, X_2, ... , X_m) = & Att(Q_1^1, K_1, V_1) \, \oplus \nonumber \\
                             & Att(Q_1^2, K_2, V_2) \oplus \cdots\\
                             & Att(Q_1^m, K_m, V_m) \nonumber
\end{align}
where $m$ is the number of modalities.

\subsection{Loss Function}
As a multi-label task, AU detection faces the common problem of data unbalances.
Hence, we apply the weighted binary cross-entropy (BCE) loss functions. 
The $AU_k$ occurrence probability function $P(AU_k)$ and the correlate positive weight $p_k$ of weighted BCE are given as:

\begin{align}
        P(AU_k) &= \frac{\sum_{i=1}^{N} y_k^i}{\sum_{i=1}^{N} \sum_{k=1}^{C} y_{k}^i}\\
        p_k &= \frac{P(AU_k)}{\min\limits_{x \in \{1,...,C\}}P(AU_x)}
\end{align}

where $N$ is the total number of training set, $C$ is the number of AU. 

The weighted BCE loss is defined:
\begin{align}
    \mathcal{L} =& - \frac{1}{N}\sum_{i=1}^{N} \sum_{k=1}^{C}(p_k\times y_k^i \times log(\widehat{y_k^i}) + \nonumber\\
       &\qquad{} (1-y_k^i) \times log(1-\widehat{y_k^i}))
\end{align}
where $y_k^i$ and $\widehat{y_k^i}$ represent the ground truth label and prediction for $AU_k$ respectively.
There are two classifiers for the two pipelines in our model, where two losses are combined.
We denote the loss for the first pipeline of fusion feature as $\mathcal{L}^{f(\alpha, \beta)}$ and the second pipeline of $\beta$ feature as $\mathcal{L}^\beta$, then the final loss can be written as:
\begin{equation}
    \mathcal{L} = \lambda_1\mathcal{L}^{f(\alpha, \beta)} + \lambda_2\mathcal{L}^{\beta}
\end{equation}
where $\lambda_1$ and $\lambda_2$ are two positive regularization parameters.

%%%%%%%%%%%%%%%%%%%%%%%%%%%%%%%%%%%%%%%%%%%%%%%%%%%%%%%%%%%%%%%%%%%%%%%%%%%%%%%%
\section{EXPERIMENTS}

To evaluate our method, we examine our framework in AU detection based on single modality and also multi-modalities.
Single modality experiments are designed to validate the efficiencies of the AU encoding by the transformer-based encoder.
We also make multi-modalities AU detection experiments for fusion quality evaluation.
We use two public multi-modal databases BP4D \cite{bp4d} and BP4D+ \cite{bp4d+} for AU occurrence detection.

\subsection{Data}

\subsubsection{BP4D}
The dataset contains 41 subjects (23 females and 18 males) captured under laboratory environments.
8 tasks are designed to elicit a range of spontaneous emotions. There are 328 = 41 × 8 sequences in total with a frame rate of 25. 
Expert coders select the most expressive 20s of each sequence for AU coding.
Around 140,000 labeled frames are included in our experiments and split into 3-fold for a fair comparison with state-of-the-art algorithms. In order to avoid training and testing on images of the same person, we use the same splits as in \cite{eac}. 
For each frame, the database provides 3D face meshes as well as corresponding texture images.

\subsubsection{BP4D+}
There are 140 subjects in BP4D+. On the basis of 8 tasks included in BP4D, two more tasks are introduced with a richer set of spontaneous emotions.
Each frame is associated with synchronized 3D face mesh, 2D texture, thermal image, and a series of physiological signals.
Similar to BP4D, 20 seconds from four tasks were AU labeled for all 140 subjects, resulting in 192,000 labeled frames.
12 Aus are selected and the performance of four-fold subject-exclusive cross-validation is reported. 

\subsection{Preprocessing}
In our experiments, we use two modalities in BP4D and BP4D+: 2D RGB image and 3D face model. 
The face regions of RGB images are detected and cropped by the publicly available library OpenCV.
A raw 3D face model contains a 30k - 50k number of vertices in BP4D and BP4D+ databases, including the face, hairs, and neck areas of a subject.
We first cropped the face area of 3D meshes and then project them into depth images.
All the face images, including RGB and Depth, are resized to 244x244 for training and testing.
We calculate the average and standard deviation of each modality in one dataset and then apply z-score standardization to each image.

\subsection{Implementation Details}
Both backbones in our model are ResNet-50 networks, which were pretrained on ImageNet \cite{imagenet}.
The layer number and head number are set 3 and 4 in the transformer encoder, respectively.
The AU embedding size is 128, the header size is 32, the MLP size is 256, and the dropout rate is 0.5.
The fusion transformer module is a one-layer and two-head transformer encoder, and the other parameters are the same as the transformer encoder except there's no dropout.
The hyper-parameter $\lambda_1$ and $\lambda_2$ are selected by comparing the best F1-score in the first fold of BP4D, which are 0.6 and 0.4 respectively. 
The comparing results is shown in Table.\ref{tab:table0}. 
We use the stochastic gradient descent (GSD) optimizer with a momentum of 0.9 and weight decay of 0.0005. 
The initial learning rate is 0.01 and 10 times smaller every epoch start from epoch 4. 
The network is trained for 5 epochs for BP4D and 6 epochs for BP4D+.
The batch size is set to 32 on both BP4d and BP4D+.
% Our framework is implemented with Pytorch \cite{pytorch} and training and testing on the GPU of NVIDIA GeForce 2080Ti.

                %%%%%% Table 0 lambda 1, lambda 2 %%%%%%

\begin{table}[th]
\caption{Hyper-parameters Comparison on one fold of BP4D}
\label{tab:table0}
\centering
\renewcommand{\arraystretch}{1.0}
\begin{tabular}{c|cccccccc}
\hline
$\lambda_1$  & 0.2  & 0.3  & 0.4  & 0.5     & 0.6   & 0.7   & 0.8  &    1.0 \\ \hline
$\lambda_2$  & 0.8  & 0.7  & 0.6  & 0.5     & 0.4   & 0.3   & 0.2  &    0.5 \\ \hline
Avg F1       & 63.1 & 63.0 & 63.5 & 64.1  & \B{64.7} & 63.7  & 63.7 &   63.5 \\ \hline
\end{tabular}
\end{table}

                %%%%%% Table 1 BP4D %%%%%%

\begin{table*}[ht]
\caption{F1 scores in terms of 12 AUs and mean accuracy on BP4D database. Bold numbers indicate the best performance in each modal; Bold-Bracketed numbers indicate the best in all modals}
\label{tab:table1}
\centering %
\renewcommand{\arraystretch}{1.0}
\begin{tabular}{l|c|*{12}{c}|c}
\hline
Methods             & Modal   & AU1       & AU2       & AU4       & AU6       & AU7       & AU10      & AU12      & AU14      & AU15      & AU17      & AU23      & AU24      & Avg\\
\hline
EAC \cite{eac}      & RGB     & 39.0      & 35.2      & 48.6      & 76.1      & 72.9      & 81.9      & 86.2      & 58.8      & 37.5      & 59.1      & 35.9      & 35.8      & 55.9\\
DSIN \cite{dsin}    & RGB     & 51.7      & 40.4      & 56.0      & 76.1      & 73.5      & 79.9      & 85.4      & 62.7      & 37.3      & 62.9      & 38.8      & 41.6      & 58.9\\
JAA \cite{jaa}      & RGB     & 47.2      & 44.0      & 54.9      & 77.5      & 74.6      & 84.0      & 86.9      & 61.9      & 43.6      & 60.3      & 42.7      & 41.9      & 60.0\\
SRERL \cite{srerl}  & RGB     & 46.9      & 45.3      & 55.6      & 77.1      & \B{78.4}  & 83.5      & 87.6      & \B{63.9}  & 52.2      & \B{63.9}  & 47.1      & 53.3      & 62.9\\ 
SEV-Net \cite{sev}  & RGB     & \BB{58.2} & \B{50.4}  & 58.3      & \B{81.9}  & 73.9      & \B{87.8}  & 87.5      & 61.6      & 52.6      & 62.2      & 44.6      & 47.6      & 63.9\\
\textbf{Ours}       & RGB     & 49.6 	  & 48.1      & \B{59.9}  & 78.4      & 78.0	  & 83.7  	  & \B{87.9}  & 62.0	  & \B{55.3}  & 61.8      & \B{50.9}  & \B{54.9}  & \B{64.2}\\
\hline
ResNet-50           & Depth   & \B{44.5}  & \B{41.3}  & \B{57.2}  & 76.5	  & 72.6	  & 80.9	  & 86.8	  & 54.3	  & \B{50.2}  & \B{62.7}  & 48.1	  & \B{44.5}  & 60.0\\ 
\textbf{Ours}       & Depth   & 38.6	  & 37.3	  & 44.2	  & \B{84.2}  & \BB{89.0} & \B{89.7}  & \BB{89.2} & \BB{79.8} & 44.7	  & 46.0	  & \BB{53.2} & 37.0	  & \B{61.1}\\ 
\hline
Late fusion         & Fusion  & 44.3	  & 34.0	  & 41.6	  & \BB{85.2} & \B{87.3}  & \BB{90.4} & 88.9      & \B{79.4}  & 46.1	  & 49.6	  & 58.6	  & 38.6	  & 62.0\\
MTUT \cite{mtut}    & Fusion  & 51.3      & 50.2      & 62.2      & 77.2      & 71.7      & 83.8      & 88.2      & 61.4      & 54.3      & 57.9      & 45.8      & 42.2      & 62.2\\ 
TEMT-Net \cite{temt}& Fusion  & \B{53.7}  & 47.1      & 60.5      & 77.6      & 75.6      & 84.8      & 87.4      & 67.0      & \BB{57.2} & 61.3      & 44.7      & 41.6      & 63.2\\ 
AMF \cite{amf}      & Fusion  & 52.1      & \BB{51.0} & \BB{64.5} & 79.2      & 73.9      & 86.4      & 88.3      & 60.5      & 55.3      & 64.2      & 47.7      & 49.2      & 64.4\\ 
\textbf{Ours}       & Fusion  & 51.6	  & 49.2	  & 57.6      & 78.8	  & 77.5      & 84.4	  & 87.9	  & 65.0	  & 56.5	  & \BB{64.3} & 49.8	  & \BB{55.1} & \BB{64.8}\\ 
\hline
\end{tabular}
\end{table*}
                %%%%%% Table 2 BP4D+ %%%%%%
\begin{table*}[ht]
\caption{F1 scores in terms of 12 AUs and mean accuracy on BP4D+ database. Bold numbers indicate the best performance in each modal; Bold-Bracketed numbers indicate the best in all modals; * indicts the result from our own implementation.}
\label{tab:table2}
\centering %
\renewcommand{\arraystretch}{1.05}
\begin{tabular}{l|c|*{12}{c}|c}
\hline
Methods             & Modal   & AU1       & AU2       & AU4       & AU6       & AU7       & AU10      & AU12      & AU14      & AU15      & AU17      & AU23      & AU24      & Avg\\
\hline
EAC \cite{eac}*     & RGB     & 43.7      & 39.0      & 14.0      & 85.6      & 87.2      & 90.5      & 88.7      & \BB{88.4} & 45.7      & 49.0      & 57.3      & \BB{43.6} & 61.1\\
JAA \cite{jaa}*     & RGB     & 46.0	  & \B{41.3}  & \B{36.0}  & 86.5	  & 88.5	  & 90.5	  & \B{89.6}  & 81.1	  & 43.4	  & 51.0	  & 56.0	  & 32.6	  & 61.9\\
SEV-Net \cite{sev}  & RGB     & 47.9      & 40.8      & 31.2      & \BB{86.9} & 87.5      & 89.7      & 88.9      & 82.6      & 39.9      & \BB{55.6} & \BB{59.4} & 27.1      & 61.5 \\
\textbf{Ours}       & RGB     & \B{48.4}  & 37.1	  & 34.4	  & 85.6	  & \BB{88.6} & \BB{90.7} & 88.8      & 81.0	  & \B{47.6}  & 51.5	  & 55.6	  & 36.9	  & \B{62.2}\\
\hline
ResNet-50           & Depth   & 35.7	  & 33.9	  & 38.9	  & 84.2	  & 86.8	  & 89.7	  & 89.1	  & 77.1	  & 40.7	  & 45.7	  & 54.4	  & 33.1	  & 59.1 \\ 
\textbf{Ours}       & Depth   & \B{43.2}  & \B{40.2}  & \B{43.3}  & \B{84.5}  & \B{88.5}  & \B{90.2}  & \B{89.3}  & \B{79.7}  & \B{45.8}  & \B{48.7}  & \B{53.8}  & \B{36.8}  & \B{62.0}\\ 
\hline
Late fusion         & Fusion  & 46.0	  & 41.3	  & 36.0      & \B{86.5}  & 88.5	  & 90.5	  & 89.6	  & 81.1	  & 43.4	  & 51.0      & 56.0	  & 32.6	  & 61.9\\ 
AMF \cite{amf}      & Fusion  & 45.3      & \BB{42.5} & 34.8      & 85.9      & 87.9      & 89.5      & \BB{90.4} & \B{82.6}  & \BB{50.1} & 45.5      & 55.7      & \B{42.1}  & 62.7\\ 
\textbf{Ours}       & Fusion  & \BB{49.6} & 42.0	  & \BB{43.5} & 85.8	  & \BB{88.6} & \B{90.6}  & 89.7	  & 80.8	  & 49.8	  & \B{52.2}  & \B{59.1}  & 38.4	  & \BB{64.2}\\ 
\hline
\end{tabular}
\end{table*}

\subsection{Results}

\subsubsection{single-modal results}

Without fusion modules, each pipeline in our network is workable for single modality AU detection. 
Hence, we first compare our method to the state-of-the-art algorithms on a RGB modality in terms of F1-score.
The methods evaluated on BP4D are including EAC \cite{eac}, DSIN \cite{dsin}, JAA \cite{jaa}, SRERL \cite{srerl}, SEV-Net \cite{sev}. 
The upper section of Table.\ref{tab:table1} shows the results on BP4D, where our method achieves the highest performance in RGB modality.
Specifically, our method outperforms all the SOTA methods in recognizing \textit{AU4}, \textit{AU12}, \textit{AU15}, \textit{AU23}, \textit{AU24}.
Our method achieves 4.2\% higher than JAA, which used facial landmarks as a joint task for AU detection.
Compare to SRERL, which requires an AU relation graph from label distribution, our method still shows 1.3\% higher performance.
Recently, SEV-Net achieved 63.9\% in F1-score by including the transformer for Intra-AU and Inter-AU attention learning from the semantic descriptions of AUs.
However, we don't need these semantic descriptions of AUs and still achieve 64.2\% by increasing 0.3\% in F1-score. 
To evaluate our method on Depth modality, we compare with ResNet-50, as the baseline function, where our method improves the performance in almost all AUs, except the \textit{AU15}, which is 0.1\% less. 
The result is reported in the middle section of Table.\ref{tab:table1}.
Our method on Depth performs better than RGB on some AUs, e.g., \textit{AU6}, \textit{AU7}, \textit{AU10}. 
Moreover, it even performs best in all modalities on the following AU: \textit{AU7}, \textit{AU12}, \textit{AU14}, \textit{AU23}.

On BP4D+, we compare with following works: EAC \cite{eac}, JAA \cite{jaa}, SEV-Net \cite{sev}. 
Since there's no report on BP4D+ in EAC and JAA, we implement these two methods for AU detection on BP4D+.
JAA implementation is based on their public source code and EAC is implemented by converting the original Theano framework to PyTorch.
The results are shown in the upper section of Table.\ref{tab:table2}.
Our method also outperforms the SOTA methods, which are 1.1\%, 1.0\%, and 0.7\% higher than EAC, JAA, and SEV-Net, respectively.
Similar to BP4D, the result of AU detection on Depth modality is shown in the middle of Table.\ref{tab:table2}, where our method is 2.9\% higher than the baseline.
The achieved F1-score at 62.0 is even higher than the other RGB-based SOTA methods, which could be caused by the high quality of the 3D model in BP4D+.
The high performance of our method proves its ability for AU encoding.

                %%%%%% Table 3 Ablation study of components %%%%%%
\begin{table*}[th]
\caption{Ablation study of effectiveness of key components of our model on BP4D database. TE represent transformer encoder; FT represent fusion transformer. Bold numbers indicate the best performance.}
\label{tab:table3}
\centering %
\renewcommand{\arraystretch}{1.0}
\begin{tabular}{l|*{12}{c}|c}
\hline
Methods                      & AU1       & AU2       & AU4       & AU6       & AU7       & AU10      & AU12      & AU14      & AU15      & AU17      & AU23      & AU24      & Avg\\
\hline
Late fusion                  & 44.3	     & 34.0	     & 41.6	     & 85.2      & 87.3      & \B{90.4}  & 88.9      & 79.4      & 46.1	     & 49.6	     & \B{58.6}	 & 38.6	     & 62.0\\ 
Late fusion + TE             & 48.0	     & 39.2	     & 37.0	     & \B{86.1}	 & 89.1	     & \B{90.4}	 & 88.7	     & \B{80.9}	 & 46.7	     & 47.8	     & 55.9	     & 39.8	     & 62.5\\ 
ResNet-50 + FT               & 47.8	     & 36.8	     & 39.7	     & 85.4	     & \B{89.2}	 & 90.3	     & \B{89.0}	 & 80.6	     & 46.5	     & 49.6	     & 58.4	     & 40.1	     & 62.8\\ 
ResNet-50 + FT + TE (Ours)   & \B{51.6}	 & \B{49.2}	 & \B{57.6}	 & 78.8	     & 77.5	     & 84.4	     & 87.9	     & 65.0	     & \B{56.5}	 & \B{64.3}	 & 49.8	     & \B{55.1}	 & \B{64.8}\\
\hline
\end{tabular}
\end{table*}

                %%%%%% Table 4 Ablation study of modality order %%%%%%
\begin{table}[th]
\caption{Ablation study of fusion modal order on BP4D and BP4D+ database.  Bold numbers indicate the best performance on each database.}
\label{tab:table4}
\centering
\renewcommand{\arraystretch}{1.0}
\begin{tabular}{c|cc|cc}
\hline
          & \multicolumn{2}{c|}{BP4D}  & \multicolumn{2}{c}{BP4D+} \\
            \cline{2-5}
          & $F(R,D)$     & $F(D,R)$    & $F(R,D)$     & $F(D,R)$\\
\hline
AU1       & \B{51.6}	 & 50.3	       & 49.3         & \B{49.6}\\ 
AU2       & \B{49.2}     & 48.9	       & 39.7         & \B{42.0}\\
AU4       & 57.6	     & \B{58.0}	   & 42.8         & \B{43.5}\\ 
AU6       & 78.8         & \B{79.6}	   & 85.4         & \B{85.8}\\ 
AU7       & \B{77.5}	 & 77.3	       & \B{88.9}     & 88.6\\ 
AU10      & 84.4         & \B{84.5}	   & \B{91.0}     & 90.6\\ 
AU12      & 87.9	     & \B{88.2}	   & \B{90.0}     & 89.7\\ 
AU14      & \B{65.0}     & 62.4        & \B{81.8}	  & 80.8\\ 
AU15      & \B{56.5}	 & 54.7	       & 47.7         & \B{49.8}\\ 
AU17      & \B{64.3}     & 61.8        & \B{53.0}	  & 52.2\\ 
AU23      & 49.8	     & \B{51.9}	   & \B{59.2}         & 59.1\\ 
AU24      & \B{55.1}     & 53.9	       & 37.5         & \B{38.4}\\ 
\hline
Avg       & \B{64.8}     & 64.3        & 63.9         & \B{64.2}\\
\hline
\end{tabular}
\end{table}

\subsubsection{multi-modal results}

Fusion refer to the fusion of RGB and Depth in the modal column of Table.\ref{tab:table1} and Table.\ref{tab:table2}.
Equation (\ref{eq:fusion}) shows $f(X_{rgb}, X_{depth})$ is not equal to $f(X_{depth}, X_{rgb})$, which means our method is sensitive to the order of fusion modalities.
Based on various experiments results of evaluating the order of modalities in fusion transformer module, we report the result of $f(X_{rgb}, X_{depth})$ on Table.\ref{tab:table1} and $f(X_{depth}, X_{rgb})$ is on Table.\ref{tab:table2}.
The differences of the fusion modalities order will be discussed in the following ablation study subsection.

Late fusion is currently the most common fusion technique, so we use late fusion with ResNet-50 as a baseline. 
Then we compare with these methods on BP4D: MTUT \cite{mtut}, TEMT-Net \cite{temt}, and AMF \cite{amf}.
The F1-score of MTUT and TEMT-Net are reported in the paper of AMF, where the author implemented them with the fusion of RGB and Depth.
Cut-switch is a data augmentation method in AMF, which is independent of AMF's deep model.
For a fair comparison, we compare AMF with its non-switched version without applying such data augmentation.
The results on BP4D are shown in the bottom section of Table.\ref{tab:table1}.
Our method outperforms all the related methods and achieves the highest F1-score at 64.8\%, which is 2.6\% higher than latefusion, 2.6\% higher than MTUT, 1.6\% higher than TEMT-Net, and 0.4\% higher than AMF.
In terms of 12 AUs, our method performs best in \textit{AU17} and \textit{AU24}.

Besides baseline, we also compare our method to AMF on BP4D+ and report the results in the bottom part of Table.\ref{tab:table2}
Our method also achieves the best performance, showing 2.3\% higher than latefusion and 1.5\% higher than AMF.
More specifically, our model achieves the best performances on \textit{AU1}, \textit{AU4}, \textit{AU7}, \textit{AU10}, \textit{AU17}, and \textit{AU23}. 

As we can see, our multi-modal method is obviously outperformed over the single modal methods on both two datasets. 
In addition, the high performance over the related multi-modal methods also verifies the effectiveness of our method.

\subsection{Ablation Study}

\subsubsection{Effect of individual part for AU detection}
For more rigorous proof of the validity of our model, we conduct experiments on BP4D dataset to examine the two major individual components, transformer encoder, and fusion transformer.
Two important components of our model, namely, transformer encoder (TE), and fusion transformer (FT), are tested as we run the same experiments using variations of the proposed network, with and without them.
Table.\ref{tab:table3} shows the performance of various combinations of the components in terms of F1-score over all the AUs.
By using the transformer encoder to train latefusion features, the performance is improved by 0.5\%, which provides another piece of evidence for the effectiveness of the transformer-based AU encoder.
By applying the fusion transformer and without the transformer encoder, the performance is 0.8\% higher than latefusion, which shows the effectiveness of the FT module.
The last row shows the proposed method, with both TE and FT, achieves the best performance. 

\subsubsection{Effect of modal fusion order}
As previously discussed, our method is affected by the order of modal fusion.
Two fusion features, $F(R, D)$ and $F(D, R)$, respectively represent the output from fusion transformer by applying  $f(X_{depth}, X_{rgb})$ and $f(X_{depth}, X_{rgb})$.
We conduct experiments on BP4D and BP4D+ datasets and report the results in Table.\ref{tab:table4}.
As we can see, $F(R, D)$ shows improved performance than $F(D, R)$ on BP4D, which is 0.5\% higher, showing fusing the projected Depth features in RGB space is more effective.
However, on BP4D+, fusion in Depth space outperforms than it in RGB space by 0.3\%.
The different quality of 3D data, different number of frames, and label imbalance may cause this difference.
It is worth noting that both of these fusion features meet or exceed the performance of the state-of-the-art methods on each database.

\subsection{Multi-modal AU Detection Analysis}

\begin{figure}[ht]
    \centering
    \includegraphics[width=0.48\textwidth]{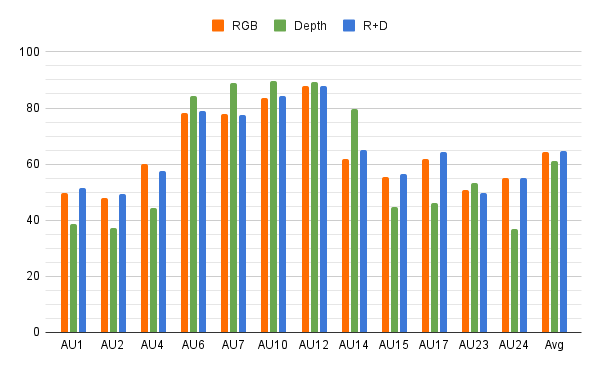}
    \caption{Performance comparison between different modalities on BP4D. R+D represents $F(R, D)$.}
    \label{fig:bp4d}
\end{figure}

\begin{figure}[ht]
    \centering
    \includegraphics[width=0.48\textwidth]{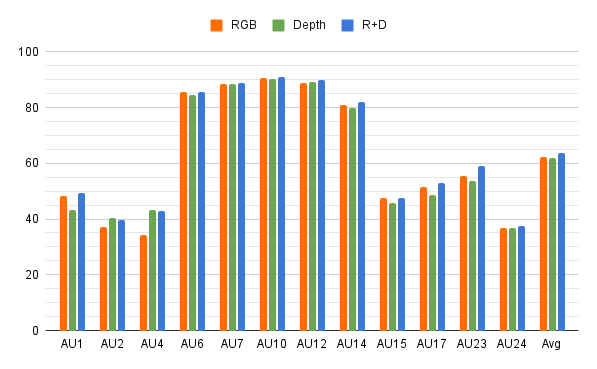}
    \caption{Performance comparison between different modalities on BP4D+. R+D represents $F(R, D)$.}
    \label{fig:bp4d+}
\end{figure}

Based on our method, we compare the recognition performance on both BP4D and BP4D+ datasets.
The result on BP4D is shown in Fig.\ref{fig:bp4d}.
Usually, the fusion modality performs better than the RGB, even when depth performs much lower.
However, the fusion result is clearly lower than the RGB on \textit{AU4}, with a 2.3\% decrease.
We propose an assumption that the Depth information plays a major role for \textit{AU4}, so the poor property of Depth modality caused performance deduction in \textit{AU4}. Fig.\ref{fig:bp4d+} shows the fusion $F(R, D)$ also performs approximately to the RGB, but more to the Depth modality on \textit{AU4} on BP4D+. 
In addition, Table.\ref{tab:table4} also shows the higher performance of $F(D, R)$ than $F(R,D)$ on \textit{AU4} of both BP4D and BP4D+. 
This also proves the depth data provided a more effective representation than the RGB on $AU4$.

Fig.\ref{fig:bp4d} shows \textit{AU6}, \textit{AU7}, \textit{AU10}, and \textit{AU14} have better performances in Depth than in RGB, which shows the high effectiveness of the AU Depth features. 
The $F(D, R)$ outperforms $F(R, D)$ in \textit{AU6}, and \textit{AU10}, but not in \textit{AU7} and \textit{AU14}, which need more exploration and discussion in our future work.

% %%%%%%%%%%%%%%%%%%%%%%%%%%%%%%%%%%%%%%%
\iffalse
from the definition: Brow Lowerer, 
Contributes to sadness, fear, and anger, and to confusion. 
Muscles: depressor glabellae, depressor supercilii, and corrugator supercilii.
% https://www.noldus.com/applications/facial-action-coding-system

• AU4: Vertical wrinkles appear in the glabella and the
    eyebrows are pulled together. The inner parts of the
    eye-brows are pulled down a trace on the right and
    slightly on the left with traces of wrinkling at the corners;

% from huiyuan CVPR2021: https://hyang428.github.io/files/CVPR21_Huiyuan_Yang-supp.pdf

\fi

%%%%%%%%%%%%%%%%%%%%%%%%%%%%%%%%%%%%%%%%%%%%%%%%%%%%%%%%%%%%%%%%%%%%%%%%%%%%%%%%
\section{CONCLUSIONS AND FUTURE WORKS}

\subsection{Conclusions}

In this paper, we have presented a novel transformer-based multi-modal network for action unit detection.
Without requiring prior knowledge like facial landmarks, AUs relation graphs, and AU semantic description, our framework utilize the transformer encoder and transformer fusion in multi-modal based AU detection. 
We evaluate our proposed method on 2 public multi-modalities databases and achieves superior performance over the peer state-of-the-art methods, not only on fusion modality but also on a single modality.
Based on the performance of our work in different modalities, we further analyze the contribution of two modalities on different AUs.

\subsection{Future Works}

Our future work will include more modalities (e.g., thermal images, physiology signals) into the fusion strategy.
Besides 3D Depth images, we also plan to develop more advanced approaches for 3D feature representations.
As we discussed previously that \textit{AU4} obtains more information from the Depth images, so we can re-design the fusion module, where different AU features are following different fusion orders.
Our work is also easy to be extended with the ROI-based attention module, which is widely used recently.
The proposed framework will also be applied to other practical affection-related applications.

%%%%%%%%%%%%%%%%%%%%%%%%%%%%%%%%%%%%%%%%%%%%%%%%%%%%%%%%%%%%%%%%%%%%%%%%%%%%%%%%
\section{ACKNOWLEDGMENTS}

The material is based on the work supported in part by the NSF under
grant CNS-1629898 and the Center of Imaging, Acoustics, and Perception
Science (CIAPS) of the Research Foundation of Binghamton University.

%%%%%%%%%%%%%%%%%%%%%%%%%%%%%%%%%%%%%%%%%%%%%%%%%%%%%%%%%%%%%%%%%%%%%%%%%%%%%%%%

{\small
\bibliographystyle{ieee}
\bibliography{egbib}

\begin{thebibliography}{10}\itemsep=-1pt

\bibitem{mtut}
M.~Abavisani, H.~R.~V. Joze, and V.~M. Patel.
\newblock Improving the performance of unimodal dynamic hand-gesture
  recognition with multimodal training.
\newblock In {\em Proceedings of the IEEE/CVF Conference on Computer Vision and
  Pattern Recognition}, pages 1165--1174, 2019.

\bibitem{layernorm}
J.~L. Ba, J.~R. Kiros, and G.~E. Hinton.
\newblock Layer normalization.
\newblock {\em arXiv preprint arXiv:1607.06450}, 2016.

\bibitem{multimodal}
T.~Baltru{\v{s}}aitis, C.~Ahuja, and L.-P. Morency.
\newblock Multimodal machine learning: A survey and taxonomy.
\newblock {\em IEEE transactions on pattern analysis and machine intelligence},
  41:423--443, 2018.

\bibitem{brown2020}
T.~B. Brown, B.~Mann, N.~Ryder, M.~Subbiah, J.~Kaplan, P.~Dhariwal,
  A.~Neelakantan, P.~Shyam, G.~Sastry, A.~Askell, et~al.
\newblock Language models are few-shot learners.
\newblock {\em arXiv preprint arXiv:2005.14165}, 2020.

\bibitem{nuscenes}
H.~Caesar, V.~Bankiti, A.~H. Lang, S.~Vora, V.~E. Liong, Q.~Xu, A.~Krishnan,
  Y.~Pan, G.~Baldan, and O.~Beijbom.
\newblock nuscenes: A multimodal dataset for autonomous driving.
\newblock In {\em Proceedings of the IEEE/CVF conference on computer vision and
  pattern recognition}, pages 11621--11631, 2020.

\bibitem{detr}
N.~Carion, F.~Massa, G.~Synnaeve, N.~Usunier, A.~Kirillov, and S.~Zagoruyko.
\newblock End-to-end object detection with transformers.
\newblock In {\em European Conference on Computer Vision}, pages 213--229,
  2020.

\bibitem{dsin}
C.~Corneanu, M.~Madadi, and S.~Escalera.
\newblock Deep structure inference network for facial action unit recognition.
\newblock In {\em Proceedings of the European Conference on Computer Vision
  ({ECCV})}, pages 309--324. Springer International Publishing, 2018.

\bibitem{cui2019}
H.~Cui, V.~Radosavljevic, F.-C. Chou, T.-H. Lin, T.~Nguyen, T.-K. Huang,
  J.~Schneider, and N.~Djuric.
\newblock Multimodal trajectory predictions for autonomous driving using deep
  convolutional networks.
\newblock In {\em 2019 International Conference on Robotics and Automation
  (ICRA)}, pages 2090--2096. IEEE, 2019.

\bibitem{bert}
J.~Devlin, M.-W. Chang, K.~Lee, and K.~Toutanova.
\newblock Bert: Pre-training of deep bidirectional transformers for language
  understanding.
\newblock In {\em NAACL-HLT (1)}, 2019.

\bibitem{vit}
A.~Dosovitskiy, L.~Beyer, A.~Kolesnikov, D.~Weissenborn, X.~Zhai,
  T.~Unterthiner, M.~Dehghani, M.~Minderer, G.~Heigold, S.~Gelly, et~al.
\newblock An image is worth 16x16 words: Transformers for image recognition at
  scale.
\newblock In {\em International Conference on Learning Representations}, 2020.

\bibitem{Ekman1997}
P.~Ekman and E.~L. Rosenberg, editors.
\newblock {\em {What the face reveals: Basic and applied studies of spontaneous
  expression using the Facial Action Coding System (FACS).}}
\newblock Oxford University Press, 1997.

\bibitem{resnet}
K.~He, X.~Zhang, S.~Ren, and J.~Sun.
\newblock Deep residual learning for image recognition.
\newblock In {\em Proceedings of the IEEE Conference on Computer Vision and
  Pattern Recognition (CVPR)}, June 2016.

\bibitem{hu2018}
Y.~Hu, M.~Modat, E.~Gibson, W.~Li, N.~Ghavami, E.~Bonmati, G.~Wang, S.~Bandula,
  C.~M. Moore, M.~Emberton, et~al.
\newblock Weakly-supervised convolutional neural networks for multimodal image
  registration.
\newblock {\em Medical image analysis}, 49:1--13, 2018.

\bibitem{Huang_2018}
X.~Huang, M.-Y. Liu, S.~Belongie, and J.~Kautz.
\newblock Multimodal unsupervised image-to-image translation.
\newblock In {\em Proceedings of the European Conference on Computer Vision
  (ECCV)}, September 2018.

\bibitem{irani2015}
R.~Irani, K.~Nasrollahi, M.~O. Simon, C.~A. Corneanu, S.~Escalera, C.~Bahnsen,
  D.~H. Lundtoft, T.~B. Moeslund, T.~L. Pedersen, M.-L. Klitgaard, et~al.
\newblock Spatiotemporal analysis of rgb-dt facial images for multimodal pain
  level recognition.
\newblock In {\em Proceedings of the IEEE Conference on Computer Vision and
  Pattern Recognition Workshops}, pages 88--95, 2015.

\bibitem{emonets}
S.~E. Kahou, X.~Bouthillier, P.~Lamblin, C.~Gulcehre, V.~Michalski, K.~Konda,
  S.~Jean, P.~Froumenty, Y.~Dauphin, N.~Boulanger-Lewandowski, et~al.
\newblock Emonets: Multimodal deep learning approaches for emotion recognition
  in video.
\newblock {\em Journal on Multimodal User Interfaces}, 10(2):99--111, 2016.

\bibitem{imagenet}
A.~Krizhevsky, I.~Sutskever, and G.~E. Hinton.
\newblock Imagenet classification with deep convolutional neural networks.
\newblock {\em Advances in neural information processing systems},
  25:1097--1105, 2012.

\bibitem{lakshminarayana2019}
N.~N. Lakshminarayana, N.~Sankaran, S.~Setlur, and V.~Govindaraju.
\newblock Multimodal deep feature aggregation for facial action unit
  recognition using visible images and physiological signals.
\newblock In {\em 2019 14th IEEE International Conference on Automatic Face \&
  Gesture Recognition (FG 2019)}, pages 1--4. IEEE, 2019.

\bibitem{srerl}
G.~Li, X.~Zhu, Y.~Zeng, Q.~Wang, and L.~Lin.
\newblock Semantic relationships guided representation learning for facial
  action unit recognition.
\newblock In {\em Proceedings of the AAAI Conference on Artificial
  Intelligence}, volume~33, pages 8594--8601, 2019.

\bibitem{li2019}
G.~Li, X.~Zhu, Y.~Zeng, Q.~Wang, and L.~Lin.
\newblock Semantic relationships guided representation learning for facial
  action unit recognition.
\newblock In {\em Proceedings of the AAAI Conference on Artificial
  Intelligence}, volume~33, pages 8594--8601, 2019.

\bibitem{li2015}
H.~Li, H.~Ding, D.~Huang, Y.~Wang, X.~Zhao, J.-M. Morvan, and L.~Chen.
\newblock An efficient multimodal 2d+ 3d feature-based approach to automatic
  facial expression recognition.
\newblock {\em Computer Vision and Image Understanding}, 140:83--92, 2015.

\bibitem{li2017}
H.~Li, J.~Sun, Z.~Xu, and L.~Chen.
\newblock Multimodal 2d+ 3d facial expression recognition with deep fusion
  convolutional neural network.
\newblock {\em IEEE Transactions on Multimedia}, 19(12):2816--2831, 2017.

\bibitem{eac}
W.~Li, F.~Abtahi, Z.~Zhu, and L.~Yin.
\newblock {EAC}-net: A region-based deep enhancing and cropping approach for
  facial action unit detection.
\newblock In {\em {IEEE} International Conference on Automatic Face {\&}
  Gesture Recognition ({FG})}, 2017.

\bibitem{temt}
P.~Liu, Z.~Zhang, H.~Yang, and L.~Yin.
\newblock Multi-modality empowered network for facial action unit detection.
\newblock In {\em 2019 IEEE Winter Conference on Applications of Computer
  Vision (WACV)}, pages 2175--2184. IEEE, 2019.

\bibitem{neverova2015}
N.~Neverova, C.~Wolf, G.~Taylor, and F.~Nebout.
\newblock Moddrop: adaptive multi-modal gesture recognition.
\newblock {\em IEEE Transactions on Pattern Analysis and Machine Intelligence},
  38(8):1692--1706, 2015.

\bibitem{niu_2019}
X.~Niu, H.~Han, S.~Yang, Y.~Huang, and S.~Shan.
\newblock Local relationship learning with person-specific shape regularization
  for facial action unit detection.
\newblock In {\em Proceedings of the {IEEE} Conference on Computer Vision and
  Pattern Recognition ({CVPR})}, 2019.

\bibitem{transfuser}
A.~Prakash, K.~Chitta, and A.~Geiger.
\newblock Multi-modal fusion transformer for end-to-end autonomous driving.
\newblock In {\em Proceedings of the IEEE/CVF Conference on Computer Vision and
  Pattern Recognition}, pages 7077--7087, 2021.

\bibitem{raffel2019}
C.~Raffel, N.~Shazeer, A.~Roberts, K.~Lee, S.~Narang, M.~Matena, Y.~Zhou,
  W.~Li, and P.~J. Liu.
\newblock Exploring the limits of transfer learning with a unified text-to-text
  transformer.
\newblock {\em arXiv preprint arXiv:1910.10683}, 2019.

\bibitem{ramachandram2017}
D.~Ramachandram and G.~W. Taylor.
\newblock Deep multimodal learning: A survey on recent advances and trends.
\newblock {\em IEEE signal processing magazine}, 34(6):96--108, 2017.

\bibitem{2019avec}
F.~Ringeval, B.~Schuller, M.~Valstar, N.~Cummins, R.~Cowie, L.~Tavabi,
  M.~Schmitt, S.~Alisamir, S.~Amiriparian, E.-M. Messner, et~al.
\newblock Avec 2019 workshop and challenge: state-of-mind, detecting depression
  with ai, and cross-cultural affect recognition.
\newblock In {\em Proceedings of the 9th International on Audio/Visual Emotion
  Challenge and Workshop}, pages 3--12, 2019.

\bibitem{jaa}
Z.~Shao, Z.~Liu, J.~Cai, and L.~Ma.
\newblock Deep adaptive attention for joint facial action unit detection and
  face alignment.
\newblock In {\em Proceedings of the European conference on computer vision
  (ECCV)}, pages 705--720, 2018.

\bibitem{mult}
Y.-H.~H. Tsai, S.~Bai, P.~P. Liang, J.~Z. Kolter, L.-P. Morency, and
  R.~Salakhutdinov.
\newblock Multimodal transformer for unaligned multimodal language sequences.
\newblock In {\em Proceedings of the conference. Association for Computational
  Linguistics. Meeting}, volume 2019, page 6558. NIH Public Access, 2019.

\bibitem{transformer}
A.~Vaswani, N.~Shazeer, N.~Parmar, J.~Uszkoreit, L.~Jones, A.~N. Gomez,
  {\L}.~Kaiser, and I.~Polosukhin.
\newblock Attention is all you need.
\newblock In {\em Advances in neural information processing systems}, pages
  5998--6008, 2017.

\bibitem{wang2012}
S.~Wang, Z.~Liu, Z.~Wang, G.~Wu, P.~Shen, S.~He, and X.~Wang.
\newblock Analyses of a multimodal spontaneous facial expression database.
\newblock {\em IEEE Transactions on Affective Computing}, 4(1):34--46, 2012.

\bibitem{yang2020}
F.~Yang, H.~Yang, J.~Fu, H.~Lu, and B.~Guo.
\newblock Learning texture transformer network for image super-resolution.
\newblock In {\em Proceedings of the IEEE/CVF Conference on Computer Vision and
  Pattern Recognition}, pages 5791--5800, 2020.

\bibitem{amf}
H.~Yang, T.~Wang, and L.~Yin.
\newblock Adaptive multimodal fusion for facial action units recognition.
\newblock In {\em Proceedings of the 28th ACM International Conference on
  Multimedia}, pages 2982--2990, 2020.

\bibitem{sev}
H.~Yang, L.~Yin, Y.~Zhou, and J.~Gu.
\newblock Exploiting semantic embedding and visual feature for facial action
  unit detection.
\newblock In {\em Proceedings of the IEEE/CVF Conference on Computer Vision and
  Pattern Recognition}, pages 10482--10491, 2021.

\bibitem{taskonomy}
A.~R. Zamir, A.~Sax, W.~Shen, L.~J. Guibas, J.~Malik, and S.~Savarese.
\newblock Taskonomy: Disentangling task transfer learning.
\newblock In {\em Proceedings of the IEEE conference on computer vision and
  pattern recognition}, pages 3712--3722, 2018.

\bibitem{bp4d}
X.~Zhang, L.~Yin, J.~F. Cohn, S.~Canavan, M.~Reale, A.~Horowitz, P.~Liu, and
  J.~M. Girard.
\newblock Bp4d-spontaneous: a high-resolution spontaneous 3d dynamic facial
  expression database.
\newblock {\em Image and Vision Computing}, 32(10):692--706, 2014.

\bibitem{bp4d+}
Z.~Zhang, J.~M. Girard, Y.~Wu, X.~Zhang, L.~Yin, et~al.
\newblock Multimodal spontaneous emotion corpus for human behavior analysis.
\newblock In {\em Proceedings of the IEEE conference on computer vision and
  pattern recognition}, pages 3438--3446, 2016.

\bibitem{zhang2018}
Z.~Zhang, S.~Zhai, L.~Yin, et~al.
\newblock Identity-based adversarial training of deep cnns for facial action
  unit recognition.
\newblock In {\em BMVC}, page 226. Newcastle, 2018.

\bibitem{drml}
K.~Zhao, W.-S. Chu, and H.~Zhang.
\newblock Deep region and multi-label learning for facial action unit
  detection.
\newblock In {\em Proceedings of the IEEE Conference on Computer Vision and
  Pattern Recognition}, pages 3391--3399, 2016.

\bibitem{zhu2020}
X.~Zhu, W.~Su, L.~Lu, B.~Li, X.~Wang, and J.~Dai.
\newblock Deformable detr: Deformable transformers for end-to-end object
  detection.
\newblock {\em arXiv preprint arXiv:2010.04159}, 2020.

\end{thebibliography}
}

\end{document}